\documentclass[runningheads]{llncs}
\usepackage[T1]{fontenc}
\usepackage{graphicx}
\usepackage{url}
\usepackage{amsmath}
\usepackage{amssymb}
\usepackage{booktabs}
\usepackage{multirow}
\begin{document}
	\title{Revisiting the Scale Loss Function and Gaussian-Shape Convolution for Infrared Small Target Detection }
	\titlerunning{Infrared Small Target Detection}
	\titlerunning{Infrared Small Target Detection}
	
	\author{Hao Li\inst{1}\thanks{Corresponding author} \and Man Fung Zhuo\inst{1}}
	\institute{
		Department of Electrical and Computer Engineering, University of Arizona, Tucson, USA\\
		\email{lihao@arizona.edu, zhuomanf@arizona.edu}
	}
	
	%
	%
	%
\maketitle              
\begin{abstract}
Infrared small target detection still faces two persistent challenges: training instability from non-monotonic scale loss functions, and inadequate spatial attention due to generic convolution kernels that ignore the physical imaging characteristics of small targets.
In this paper, we revisit both aspects.
For the loss side, we propose a \emph{diff-based scale loss} that weights predictions according to the signed area difference between the predicted mask and the ground truth, yielding strictly monotonic gradients and stable convergence. We further analyze a family of four scale loss variants to understand how their geometric properties affect detection behavior.
For the spatial side, we introduce \emph{Gaussian-shaped convolution} with a learnable scale parameter to match the center-concentrated intensity profile of infrared small targets, and augment it with a \emph{rotated pinwheel mask} that adaptively aligns the kernel with target orientation via a straight-through estimator.
Extensive experiments on IRSTD-1k, NUDT-SIRST, and SIRST-UAVB demonstrate consistent improvements in $mIoU$, $P_d$, and $F_a$ over state-of-the-art methods. We release our anonymous code and pretrained models.

\end{abstract}
\section{Introduction}
Infrared small target detection (IRSTD) is a vision task used in surveillance, early warning, search and tracking, and autonomous guidance in civilian and military applications. Unlike conventional object detection, IRSTD targets occupy only a few pixels in the image and typically have low signal-to-noise ratio (SNR), low signal-to-noise ratio (SNR), and limited structural or semantic information\cite{kou2023infrared,zhao2022single}. 

Traditional infrared small target detection methods can be categorized into three major paradigms: filtering-based\cite{bae2010small,bai2011hit,deng2018adaptive}, local-contrast-based\cite{chen2013local,zhang2016infrared,zhang2020small,aghaziyarati2019small}, and low-rank-based\cite{zhu2020tnlrs,zhang2019infrared,zhang2018infrared} approaches. Filtering-based methods employ filtering to enhance target-background contrast, while contrast-based methods rely on handcrafted operations to highlight local target saliency; low-rank-based methods further utilize matrix decomposition techniques to separate sparse small targets from cluttered backgrounds. Despite their intuitive design, these traditional techniques suffer from critical limitations: they heavily depend on manually engineered features rather than data-driven representations, lack adaptive modeling for complex real-world backgrounds, and often produce high false alarm rates when facing extremely small, low signal-to-noise ratio (SNR) infrared targets.
\begin{figure}[ht!]
	\centering
	\includegraphics[width=1.0\textwidth]{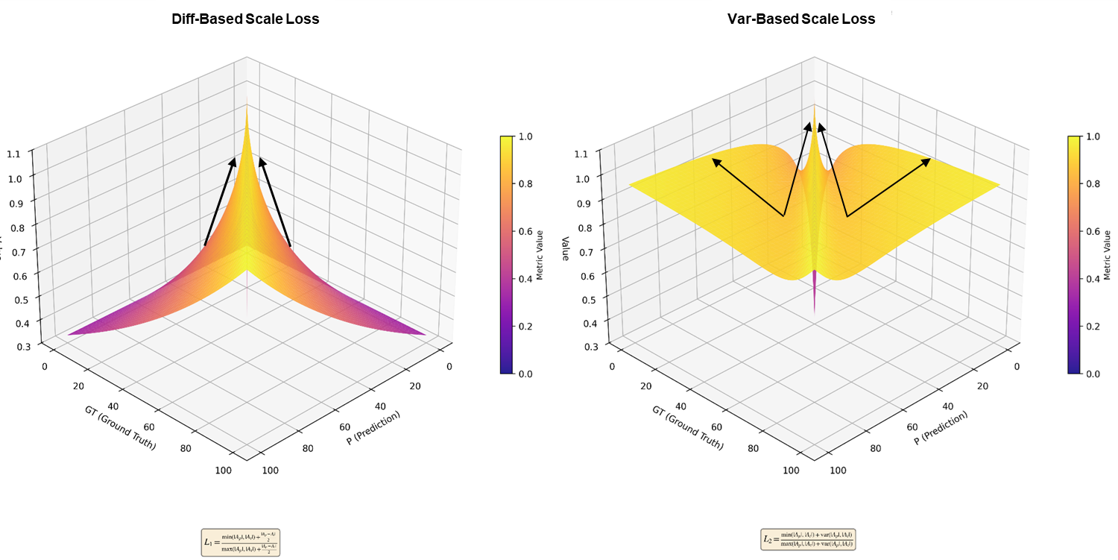} 
	\caption{Comparison between our monotonic diff-based scale loss and conventional non-monotonic scale losses. The proposed diff-based loss exhibits strict monotonicity with respect to scale deviation, stably penalizing mismatches between predicted and ground-truth target areas and naturally guiding optimization toward the optimal center, thus eliminating unstable gradients and ensuring stable convergence during training. In contrast, the var-based scale loss exhibits non-monotonic behavior, creating multiple local optima and erratic gradients that hinder stable convergence to the correct target scale.}
	\label{fig:scale_loss}
\end{figure}

\begin{figure}[h]
	\centering
	\includegraphics[width=1.0\textwidth]{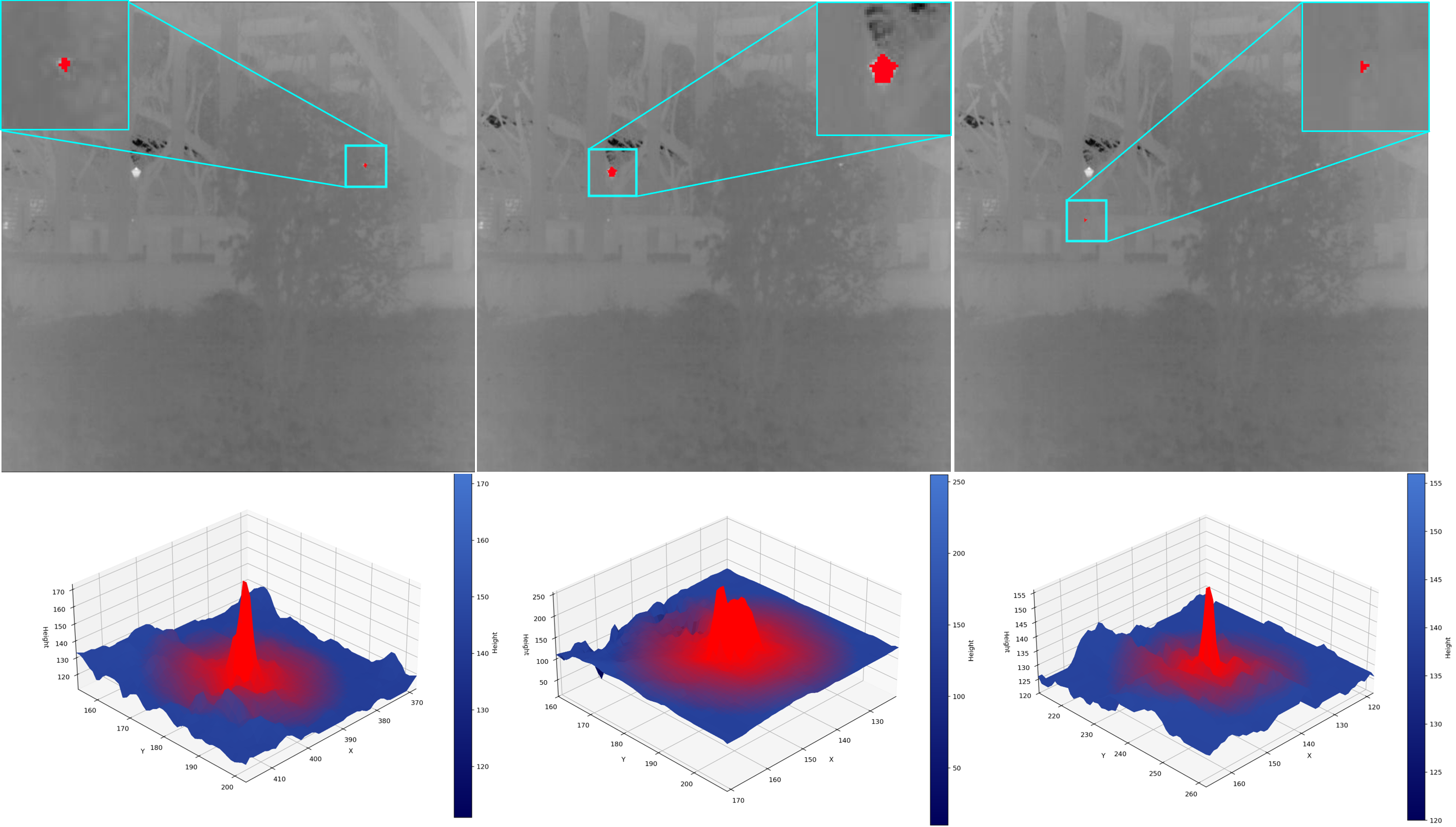} 
	\caption{Illustration of the Gaussian-like intensity distribution inherent to IRSTs. As shown, small targets exhibit a distinct center-concentrated, smoothly decaying grayscale pattern, which motivates the design of Gaussian-shaped spatial attention or convolution to align with this natural imaging characteristic, rather than using generic fully learned receptive fields.}
	\label{fig:gaussian_conv}
\end{figure}

\begin{figure}[h]
	\centering
	\includegraphics[width=1.0\textwidth]{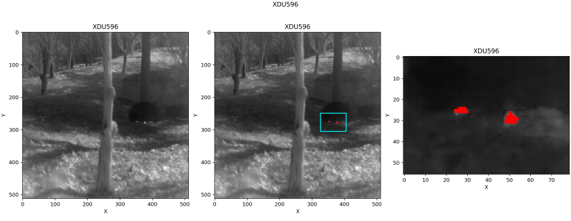} 
	\caption{Illustration of the directional morphological diversity of IRSTs. As shown, IRSTs exhibit varied directional structures (horizontal, vertical, or oblique point-like/elongated shapes), which motivates the use of learnable rotated pinwheel masks to adaptively align spatial attention with target orientation, complementing Gaussian-shaped convolution to better match the diverse directional properties of small targets.
	}
	\label{fig:pinwheel_mask}
\end{figure}

Recent deep learning methods have demonstrated superior
adaptability and accuracy by leveraging large-scale dataset\cite{dai2021asymmetric,wang2022interior,zhang2022exploring}, spatical and channel attention\cite{li2022dense,wu2022uiu,yuan2024sctransnet}, training-objective loss function design\cite{zhang2022isnet,liu2024infrared}.  However, most current CNN-based frameworks focus on complex architectural designs, yet two critical limitations remain unaddressed: non-monotonic scale loss behavior and unstructured spatial attention. 
First, common losses such as cross-entropy or standard IoU do not explicitly couple the scale of the prediction to that of the target. When the network over- or under-segments, the total predicted area can differ greatly from the ground-truth area, therefore a scale-aware term\cite{liu2024infrared} can penalize this mismatch and stabilize training. 
We revisit the scale loss in SLS and find that function exhibit non-monotonic behavior, which will hinder stable convergence to the correct target scale (as Fig.\ref{fig:scale_loss} shown). Based on this finding, we propose a diff-based scale loss which
 is built from the difference between the total predicted and target activation and is used to reweight the IoU term. 

Second, spatial attention in most models is implemented via simple fully learned convolutions, which tend to distribute weights uniformly and easily introduce false alarms in cluttered backgrounds. As illustrated in Fig.\ref{fig:gaussian_conv}, infrared small targets exhibit a distinct Gaussian-like intensity distribution—with peak values concentrated at the center and decaying smoothly toward the edges—motivating us to revisit  Gaussian-shaped convolution: a $7\times 7$ kernel with a learnable Gaussian parameter $\sigma$, enabling the receptive field to adaptively match the center-concentrated intensity distribution of infrared small targets. Furthermore, as shown in Fig.\ref{fig:pinwheel_mask}, infrared small targets often appear as diverse directional structures (horizontal, vertical, or oblique point-like/elongated shapes), so we modulate the Gaussian kernel using a pinwheel mask with learnable rotated mask to concentrate spatial attention along a narrow directional strip passing through the center, aligning with the inherent morphological properties of small targets. Together, the learnable Gaussian kernel and rotatable pinwheel mask provide a concise and interpretable spatial prior while maintaining end-to-end trainability.

In this paper, we unify the formulation of the diff-based scale loss, the Gaussian-shaped convolution with learnable $\sigma$, and the learnable rotated pinwheel mask. We integrate these components into an encoder-decoder network equipped with channel and spatial attention, and conduct extensive evaluations on mainstream infrared small target detection benchmarks. On IRSTD-1k, our best configuration achieves 69.19\% mIoU, 94.22\% $P_d$, and $7.67\times10^{-6}$ $F_a$, surpassing prior state-of-the-art across all three metrics. Consistent gains are observed on NUDT-SIRST and SIRST-UAVB. Our anonymous code and pretrained models are publicly available.\footnote{\url{https://anonymous.4open.science/r/Revisiting_the_Scale_Loss_Function_and_Gaussian_Shape_Convolution-7572}}

\section{Related Works}
\subsection{Single-Frame Infrared Small Target Detection}
Early works such as MDvsFA cGan\cite{wang2019miss} and TBC-Net \cite{zhao2019tbc} introduced convolutional neural networks to model target-background separation, yet at that time there lacked a unified, standardized benchmark for fair evaluation. Subsequently, ACM\cite{dai2021asymmetric} not only pioneered shape-aware modeling via point-wise convolution to capture the compact morphology of small targets, but also released the first standardized infrared small target dataset SIRST, which has since become the de facto benchmark for evaluating IRSTD methods. By explicitly modeling target shape and providing a unified evaluation protocol, ACM effectively suppressed background clutter, reduced false alarms, and laid the foundation for subsequent data-driven and shape-aware IRSTD research. ALCNet \cite{dai2021attentional} extended this paradigm by integrating local-contrast convolution into the ACM framework to further enhance target saliency. EAAU-Net\cite{tong2021eaau} proposed an enhanced asymmetric attention U-Net to preserve tiny target features in deep layers by designing an enhanced asymmetric attention module for same-layer feature interaction and cross-layer fusion.

The weak discriminability and tiny size of infrared small targets make it difficult for standard losses to effectively constrain target localization and separation, often leading to blurred contours and unstable training. ISNet \cite{zhang2022isnet} proposed asymmetric convolution and a composite Dice and Edge loss to explicitly model target boundary information, demonstrating that edge-sensitive constraints could refine target contours and improve detection precision. Subsequent works including IAANet \cite{wang2022interior} and FC3-Net \cite{zhang2022exploring} further deepened the research on loss functions, exploring more effective constraint strategies to optimize target localization and suppress background clutter.   In contrast, later studies shifted their focus to backbone network improvement to enhance feature representation capability. DNANet\cite{li2022dense} introduced a dense nested attention network to strengthen the backbone’s ability to capture multi-scale spatial dependencies, MTUNet\cite{wu2023mtu}  designed a cross-scale feature fusion module to optimize the backbone’s multi-resolution feature alignment, UIU-Net\cite{wu2022uiu} and SCTransNet\cite{yuan2024sctransnet} further improved backbone performance through interactive fusion and transformer integration.

\subsection{Spatical Attention for IRSTD}

Early spatial attention designs for IRSTD primarily leveraged handcrafted or fixed priors to enhance local contrast. For instance, ACM\cite{dai2021asymmetric} introduced point-wise convolution to model the compact shape of small targets, concentrating attention on local high-contrast regions and laying the foundation for shape-aware spatial attention. ALCNet \cite{dai2021attentional} extended this paradigm by integrating local-contrast convolution, further strengthening the network’s focus on target-like regions. IAANet\cite{wang2022interior} pioneered the use of transformer encoders to model pixel-level interior attention within coarse target regions, enabling the network to capture the inherent correlation between target pixels and distinguish them from background clutter. Recent advances have shifted toward adaptive spatial attention mechanisms that incorporate target-specific spatial priors, with Gaussian-shaped and pinwheel-shaped designs emerging as effective paradigms. Gaussian-shaped spatial attention draws inspiration from the natural intensity distribution of infrared small targets—where target pixels exhibit a Gaussian-like grayscale profile—and uses a kernel with learnable scale parameters to adaptively adjust the receptive field. Xu et al. \cite{xu2025small} emphasized that real infrared dim targets exhibit strong Similarity of Gaussian  in local grayscale distribution, while edge clutter and structured noise do not conform to Gaussian distribution. Their method uses Gaussian similarity as spatial prior to filter suspicious points, guiding spatial attention to retain Gaussian-like target regions while suppressing non-Gaussian clutter. Zhang et al. \cite{zhang2025gaussian} proposed a Gaussian weighted multi-scale spatial attention mechanism, which computes local correlation between image patches and 2D Gaussian models to weight multi-scale features. Yang et al. \cite{yang2025pinwheel} further extended this line by constructing a directional, center-weighted receptive field that mimics both Gaussian attenuation and directional diffusion of real targets. It expands the receptive field efficiently while concentrating weights along central strips, matching both Gaussian intensity and directional structure of dim small targets.
Zhang et al. \cite{zhang2025infrared} also exploited Gaussian-like grayscale homogeneity in their interpretation weighted sparse framework, using structure tensor analysis to enhance dim targets and using spatial contrast prior to suppress residual clutter.

\subsection{Loss Function for IRSTD}
Prior work on IoU-based objectives in generic object detection further suggests that the effectiveness of such losses depends not only on overlap itself, but also on how geometric discrepancies are parameterized and weighted. GIoU mitigates optimization saturation in non-overlapping cases\cite{rezatofighi2019generalized}. DIoU and CIoU introduce penalties on center distance and aspectratio discrepancy\cite{zheng2020distance}. Alpha-IoU shows that modifying the curvature of the loss redistributes gradients and affects localization accuracy\cite{he2021alpha}. Infrared small targets may span only a handful of pixels, and minimal deviations in area or centroid can materially affect detection probability and false-alarm behavior. This motivates a re-examination of area weighting in IRSTD-specific loss design.  ISNet\cite{zhang2022isnet} addresses the blurred segmentation caused by conventional losses failing to capture target shape and boundaries by using a composite loss consisting of Dice loss for alleviating foreground–background imbalance and a weighted combination of Dice and BCE loss for explicit edge supervision. To further enhance scale and location awareness, MSHNet\cite{liu2024infrared} proposes Scale and Location Sensitive loss, which dynamically weights IoU according to target scale and introduces polar-coordinate location constraints to improve regression precision. Building upon SLS loss, the pinwheel convolution framework\cite{yang2025pinwheel} further proposes Scale-based Dynamic loss, which adaptively adjusts the contributions of scale loss and location loss based on target size, leading to more stable optimization and better detection performance on dim and tiny targets.

\section{Methodologies}

\begin{figure}[t]
	\centering
	\includegraphics[width=1.0\textwidth]{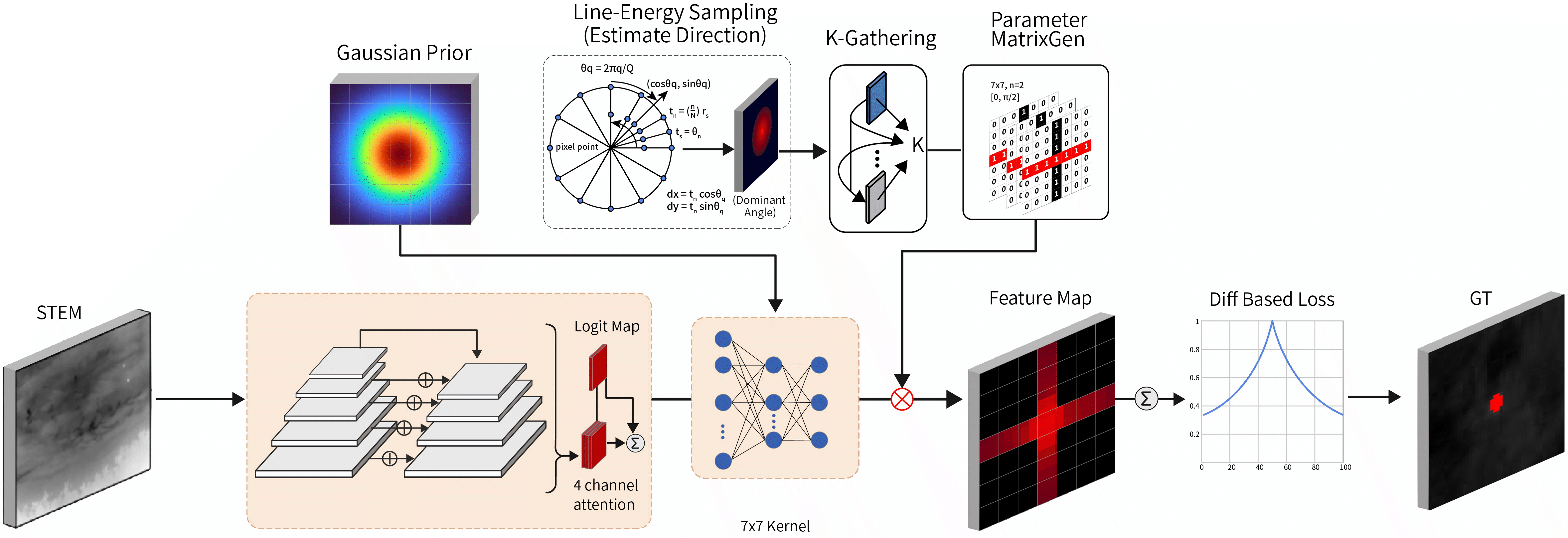}
	\caption{Overview of the proposed framework. The U-Net encoder–decoder backbone applies channel attention and Gaussian-shaped spatial attention within each residual block. The $7\times7$ spatial attention kernel is constructed by combining a Gaussian prior (learnable $\sigma$) with a learnable rotated pinwheel mask whose orientation $\theta$ is optimized via a straight-through estimator, as illustrated in the upper branch. The final prediction is supervised by the diff-based scale loss against the ground-truth mask.}
	\label{fig:overview}
\end{figure}

Fig.~\ref{fig:overview} illustrates the overall architecture of the proposed framework. The backbone is a U-Net-style encoder–decoder with four encoding stages and four decoding stages connected via skip connections. Each residual block is equipped with \emph{channel attention}—recalibrating channel-wise responses via average- and max-pooling followed by a shared MLP—and \emph{Gaussian-shaped spatial attention}, which constitutes the core of our spatial prior. 
Within each spatial attention module, the channel-attended feature map is compressed into a 2-channel descriptor via channel-wise average and max pooling, and then convolved with a $7\times7$ kernel constructed as follows. A Gaussian prior with learnable scale $\sigma$ provides the base kernel shape, capturing the center-concentrated intensity distribution of infrared small targets. 
To encode the \emph{directional} structure of targets, candidate orientations are sampled uniformly over $[0, \pi)$, and for each orientation the corresponding line-energy response is estimated; the dominant direction is selected and used to parameterize a binary $7\times7$ pinwheel mask—a disk-constrained directional strip centered at the kernel origin—via the learned rotation angle $\theta$. The final $7\times7$ kernel is the element-wise product of $\tilde{G}(\sigma)$ and this binary mask, normalized to unit sum. The resulting spatial attention map is multiplied back onto the feature map to concentrate detection on target-like regions.

\subsection{Diff-based Scale Loss Function}

\subsubsection{Notation and IoU term.}
Let $P = \sigma(\hat{y})$ be the sigmoid prediction and $G$ the binary ground-truth mask. Define the soft areas $A_p = \sum P$, $A_t = \sum G$, and the intersection $S = \sum(P \cdot G)$. The standard soft IoU is
\begin{equation}
\text{IoU} = \frac{S + \epsilon}{A_p + A_t - S + \epsilon},
\end{equation}
where $\epsilon$ is a small constant for numerical stability.

\subsubsection{L1 scale factor $\alpha$.}
Let $\Delta = |A_p - A_t|$ denote the absolute area discrepancy. We define the diff-based scale factor as
\begin{equation}
\alpha = \frac{\min(A_p, A_t) + \Delta/2}{\max(A_p, A_t) + \Delta/2}.
\end{equation}
By construction, $\alpha = 1$ when $A_p = A_t$, and $\alpha$ decreases \emph{strictly and monotonically} toward zero as $\Delta$ grows. This monotonicity is the key property that distinguishes our formulation: gradient signals consistently and stably push the network toward the correct target scale, eliminating the oscillatory behavior of the variance-based loss.

\subsubsection{Full loss and training schedule.}
Let $\mathcal{L}_{\mathrm{loc}}$ denote an optional location regularizer (detailed in Sec.~\ref{sec:weightediou}). To stabilize early training, we employ a warm-up schedule: for the first $E_{\mathrm{warm}}$ epochs we use only the standard IoU loss $\mathcal{L} = 1 - \text{IoU}$ to allow the network to develop a coarse segmentation; thereafter, we activate the full diff-based scale loss
\begin{equation}
\mathcal{L} = 1 - \alpha \cdot \text{IoU} + \mathcal{L}_{\mathrm{loc}}.
\end{equation}
The factor $\alpha$ directly couples scale consistency into the training objective: when the predicted area matches the target, $\alpha\approx 1$ and the loss reduces to the standard IoU loss; as the scale mismatch grows, $\alpha$ suppresses the IoU reward, steering optimization toward both accurate overlap and correct scale simultaneously.

\subsection{Gaussian-shaped Convolution}
\label{sec:gaussian}

\subsubsection{Gaussian kernel on a $7\times7$ grid.}
Standard spatial attention relies on a fully learned $7\times7$ convolution that imposes no structural prior, and therefore fails to exploit the well-documented center-bright, radially decaying intensity profile of infrared small targets (Fig.~\ref{fig:gaussian_conv}). We replace it with a kernel whose weights follow an isotropic Gaussian. Indexing spatial positions by offsets $i,j\in\{-3,\ldots,3\}$ from the center, the unnormalized weights are
\begin{equation}
\tilde{G}(i,j;\sigma) = \exp\!\left(-\frac{i^2+j^2}{2\sigma^2+\epsilon}\right),
\end{equation}
where $\epsilon=10^{-8}$ ensures numerical stability. The scale $\sigma$ is \emph{learnable per attention layer}: we store $\log\sigma$ as a trainable scalar, recover $\sigma=\exp(\log\sigma)$, and clamp $\sigma\in[10^{-2},10]$ to keep the kernel non-degenerate. 

\subsubsection{Masking and normalization.}
The isotropic Gaussian is insensitive to target orientation. We modulate it element-wise by a $7\times7$ binary directional mask $M$ (derived in Sec.~\ref{sec:pinwheel}), then normalize to unit sum:
\begin{equation}
K(i,j) = \frac{\tilde{G}(i,j;\sigma)\cdot M(i,j)}{\displaystyle\sum_{i',j'}\tilde{G}(i',j';\sigma)\cdot M(i',j')+\epsilon}.
\end{equation}
Normalization preserves the global activation scale of the feature map, while the product $\tilde{G}\cdot M$ encodes both the Gaussian intensity prior and the target's dominant orientation in a single compact kernel.


\subsection{Rotated Pinwheel Mask}
\label{sec:pinwheel}

\subsubsection{Geometry of the pinwheel mask.}
The isotropic Gaussian kernel encodes the radial intensity prior of small targets but cannot capture their \emph{directional} morphology. As shown in Fig.~\ref{fig:pinwheel_mask}, infrared small targets frequently exhibit elongated or asymmetric structures along a dominant orientation. We therefore introduce a \emph{learnable rotated pinwheel mask}: a $7\times7$ binary pattern shaped as a narrow directional strip intersected with a disk, whose orientation $\theta$ is optimized end-to-end. Using coordinates $(u,v)\in\{-3,\ldots,3\}^2$ relative to the center, let $r=\sqrt{u^2+v^2}$. For orientation angle $\theta$, the perpendicular distance from $(u,v)$ to the line through the origin with direction $\theta$ is
\begin{equation}
d(u,v;\theta) = -\sin\theta\cdot u + \cos\theta\cdot v.
\end{equation}
With disk radius $m=3$ and strip half-width $\mathit{tol}=0.5$, the pinwheel is the set of pixels satisfying $r\le m$ and $|d|\le\mathit{tol}$, with the center $(0,0)$ always included. This design concentrates the receptive field along a single directional strip, aligning spatial attention with the elongated morphology of real infrared small targets.

\subsubsection{Differentiable soft mask.}
Hard thresholding the disk and strip constraints blocks gradient flow to $\theta$. We replace them with sigmoid approximations controlled by a learnable sharpness parameter $\tau>0$:
\begin{align}
D(u,v) &= \sigma_s\!\left(\tfrac{m-r}{\tau}\right), \\
L(u,v;\theta) &= \sigma_s\!\left(\tfrac{\mathit{tol}-|d(u,v;\theta)|}{\tau}\right),
\end{align}
where $\sigma_s$ denotes the sigmoid function. The soft pinwheel mask is
\begin{equation}
M_{\mathrm{soft}}(u,v;\theta,\tau) = D(u,v)\cdot L(u,v;\theta),
\end{equation}
with center pixel $(0,0)$ forced to 1. The orientation is parameterized as $\theta=\theta_{\mathrm{init}}+\theta_{\mathrm{rot}}$, initialized to $0$ and $\pi/6$ respectively, both learnable.

\subsubsection{Straight-through estimator.}
A binary mask in the forward pass is essential to maintain a sparse, interpretable receptive field and to ensure stable normalization of the Gaussian kernel; yet hard binarization blocks gradient flow to $\theta$ and $\tau$. We resolve this tension via the straight-through estimator: let $\widehat{M}=\mathbb{1}[M_{\mathrm{soft}}>0.5]$ be the binary mask, and define
\begin{equation}
M_{\mathrm{out}} = \widehat{M} + \bigl(M_{\mathrm{soft}} - \mathrm{sg}(M_{\mathrm{soft}})\bigr),
\end{equation}
where $\mathrm{sg}(\cdot)$ denotes the stop-gradient operator. In the forward pass $M_{\mathrm{out}}=\widehat{M}$ (binary), preserving kernel sparsity; in the backward pass gradients flow through $M_{\mathrm{soft}}$ to update $\theta$ and $\tau$, enabling continuous orientation adaptation. The final spatial attention kernel is $K=\mathrm{normalize}\bigl(\tilde{G}(\sigma)\cdot M_{\mathrm{out}}\bigr)$ as defined in Sec.~\ref{sec:gaussian}.

\section{Experiments}
\subsection{Experiment Setup}

\subsubsection{Environments.}
All experiments are conducted under the PyTorch framework. Exploratory and ablation studies are prototyped on a local workstation equipped with a single NVIDIA GeForce RTX~5080 GPU (16\,GB GDDR7, CUDA~13.1, Driver~591.55). Large-scale comparative experiments are performed on a high-performance computing node equipped with a single NVIDIA GH200~GPU (96\,GB HBM3e, CUDA~12.8, Driver~570.195.03), which provides sufficient memory capacity for training with larger batch sizes and higher-resolution inputs.
\subsubsection{Implementation Details.}
We conduct experiments on three benchmarks. \textbf{IRSTD-1k}~\cite{zhang2022isnet} contains 1,000 infrared images at $512\times512$ resolution with relatively larger single-frame targets across diverse real-world scenes. \textbf{NUDT-SIRST}~\cite{li2022dense} is a synthetic dataset of 1,327 clipped infrared image patches at $256\times256$ resolution, covering ten scene categories (sky, sea surface, ground clutter, etc.) with high intra-class diversity and densely annotated small targets. \textbf{SIRST-UAVB}~\cite{yang2025pinwheel} comprises 3,000 images capturing small UAVs and birds in complex real-world backgrounds, posing additional challenges from target morphology variation and cluttered scenes.
All training settings are the same to maintain consistency: 400 epochs, Adagrad optimizer with learning rate 0.05, batch size 4, and input resolution $256 \times 256$. Model selection is based on the best validation mIoU, with the corresponding checkpoint saved.
\subsubsection{Datasets and Metrics.}
We evaluate on mainstream IRSTD benchmarks. All three metrics are computed on binarized predictions $\hat{y} = \mathbf{1}[\text{logits} > 0]$ (i.e., sigmoid output $> 0.5$).

\textbf{mIoU} (mean Intersection over Union) is computed at the pixel level over the full validation set:
\begin{equation}
\text{IoU} = \frac{|\hat{Y}\cap Y|}{|\hat{Y}\cup Y|}, \quad \text{mIoU} = \text{mean}(\text{IoU}),
\end{equation}
where $\hat{Y}$ and $Y$ denote the predicted and ground-truth foreground pixel sets; the metric is accumulated over all images and then averaged.

\textbf{Pd} (probability of detection) and \textbf{Fa} (false alarm rate) are evaluated at the object level. Connected components are extracted from both the binarized prediction and the ground-truth mask using 8-connectivity. A predicted component is counted as a true detection if its centroid lies within 3 pixels of a ground-truth centroid; once matched, the predicted component is removed so that each predicted component is matched at most once. Fa accumulates the total pixel area of all unmatched predicted components and is reported in units of $\times 10^6$:
\begin{equation}
\text{Pd} = \frac{\#\,\text{detected targets}}{\#\,\text{ground-truth targets}},
\end{equation}
\begin{equation}
\text{Fa} = \frac{\sum\text{area of unmatched predicted components}}{N \times 256 \times 256} \times 10^6,
\end{equation}
where $N$ is the number of validation images.

\begin{figure}[t]
	\centering
	\includegraphics[width=1.0\textwidth]{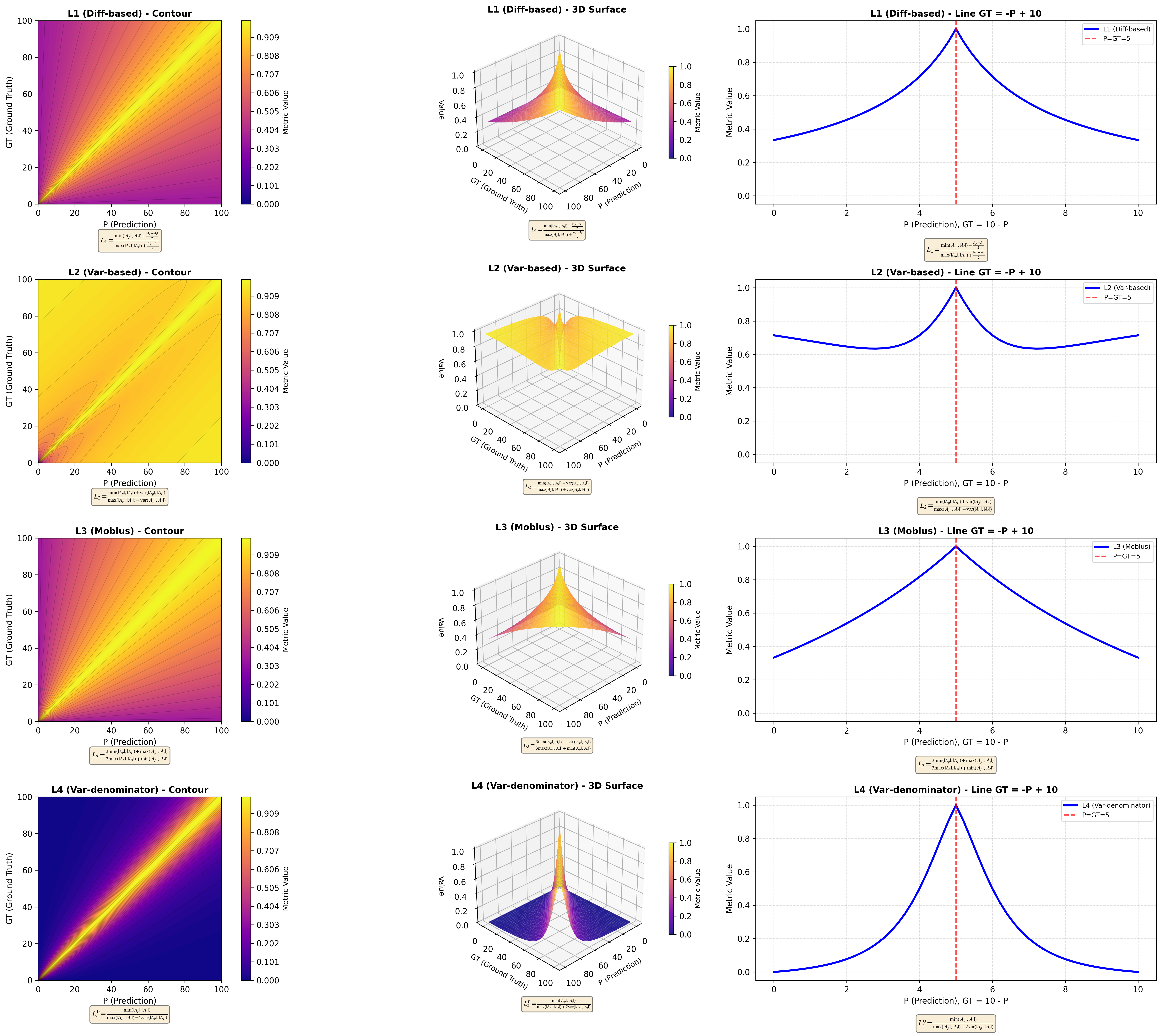}
	\caption{Visualization of the four scale weighting functions. Each row corresponds to one variant (Diff-based, Var-based, Mobius, Var-denominator). \textbf{Left}: contour map over the $(A_p, A_t)$ plane. \textbf{Middle}: 3D surface. \textbf{Right}: anti-diagonal cross-section with $A_p + A_t = 10$, showing the weight value as $A_p$ varies. Only the Diff-based weight decays strictly and monotonically away from $A_p = A_t$, while the Var-based weight is non-monotonic.}
	\label{fig:loss_family}
\end{figure}

\subsection{Ablation Study}
\subsubsection{Scale Loss Family}
We define the standard soft IoU as the base overlap term, where \(I = \sum (\mathbf{P} \odot \mathbf{G})\) is the soft intersection, \(A_p = \sum \mathbf{P}\) and \(A_t = \sum \mathbf{G}\) are the soft areas of prediction and ground truth:
\[
J(\mathbf{P}, \mathbf{G}) = \frac{I + \epsilon}{A_p + A_t - I + \epsilon}.
\]
To incorporate scale sensitivity, we introduce a family of scale weighting functions \(w(A_p, A_t)\) that modulate the IoU term. All weights satisfy \(w=1\) when \(A_p=A_t\). Let \(\Delta = |A_p - A_t|\), \(m = \min(A_p, A_t)\), and \(M = \max(A_p, A_t)\). We benchmark four profiles: our proposed Diff-based loss, the original Var-based loss from~\cite{liu2024infrared}, and two additional novel variants (Mobius and Var-denominator) we design to span a wider spectrum of scale sensitivity, enabling a systematic comparison of monotonic vs.\ non-monotonic and mild vs.\ aggressive penalization behaviors.

\textbf{Diff-based} (proposed): uses the L1 absolute difference, yielding \emph{strict monotonic} decay:
\[
w_{\text{Diff}} = \frac{m + \Delta/2}{M + \Delta/2 + \epsilon}.
\]

\textbf{Var-based} (baseline from~\cite{liu2024infrared}): uses the squared half-difference, resulting in \emph{non-monotonic} behavior where large deviations are under-penalized:
\[
w_{\text{Var}} = \frac{m + \left(\Delta/2\right)^2}{M + \left(\Delta/2\right)^2 + \epsilon}.
\]

\textbf{Mobius} (proposed for comparison): a ratio-based formulation whose value naturally lies in $[1/3,\,1]$, offering a moderate and smooth monotonic response with a higher weight baseline:
\[
w_{\text{Mob}} = \frac{3m + M}{3M + m + \epsilon}.
\]

\textbf{Var-denominator} (proposed for comparison): places the squared discrepancy only in the denominator, imposing the strongest monotonic penalty and concentrating weight in a narrow region around $A_p = A_t$:
\[
w_{\text{VD}} = \frac{m}{M + 2\left(\Delta/2\right)^2 + \epsilon}.
\]

All four weights are integrated into the loss as \(\mathcal{L} = 1 - w\cdot J + \mathcal{L}_{\text{loc}}\), as described in Sec.~\ref{sec:weightediou}.

Fig.~\ref{fig:loss_family} visualizes the four weighting functions as contour maps, 3D surfaces, and anti-diagonal cross-sections (fixing $A_p + A_t = 10$). The contour and 3D plots reveal that $w_{\text{Diff}}$ decays smoothly and strictly as the prediction-target discrepancy grows, while $w_{\text{Var}}$ exhibits a non-monotonic ``valley-ridge'' shape—weights recover as the area gap becomes very large, creating spurious gradient directions. $w_{\text{Mob}}$ is smoother and less aggressive, maintaining a relatively high value across a wider region. $w_{\text{VD}}$ concentrates nearly all weight near the diagonal and penalizes mismatches most aggressively, effectively acting as a spike. 
The remaining three variants all decrease monotonically, with $w_{\text{Diff}}$ exhibiting a smooth and moderate slope, $w_{\text{Mob}}$ maintaining a higher baseline, and $w_{\text{VD}}$ concentrating weight in an extremely narrow region around the optimum.

\subsubsection{Location Loss}
\label{sec:weightediou}
We employ a location regularizer \(\mathcal{L}_{\text{loc}}\) based on the centroids of \(\mathbf{P}\) and \(\mathbf{G}\). Let \((\bar{x}_p, \bar{y}_p)\) and \((\bar{x}_t, \bar{y}_t)\) be the centroids in normalized coordinates. The regularizer penalizes discrepancies in both angle and radial distance:
\[
\mathcal{L}_{\text{loc}} = \frac{4}{\pi^2}\left( \arctan\frac{\bar{y}_p}{\bar{x}_p} - \arctan\frac{\bar{y}_t}{\bar{x}_t} \right)^2 + \left( 1 - \frac{\min(\rho_p, \rho_t)}{\max(\rho_p, \rho_t)} \right)
\]
where \(\rho_p = \sqrt{\bar{x}_p^2 + \bar{y}_p^2}\) and \(\rho_t = \sqrt{\bar{x}_t^2 + \bar{y}_t^2}\). 

\subsubsection{Quantitive Results}

Table~\ref{tab:unified_compare_fused} reports mIoU, $P_d$, and $F_a$ on three standard datasets: IRSTD-1k, NUDT-SIRST, and SIRST-UAVB. Method names follow the format \texttt{L\{1--4\}-[GP][-Rotated/ONLY]}, where \texttt{L1}--\texttt{L4} denote the scale loss variant (specifically, \texttt{L1}: Diff-based, \texttt{L2}: Var-based, \texttt{L3}: Mobius, \texttt{L4}: Var-denominator), \texttt{GP} indicates Gaussian-shaped spatial attention, \texttt{Rotated} adds the learnable rotated pinwheel mask, and \texttt{ONLY} omits the location regularizer $\mathcal{L}_{\text{loc}}$.

On IRSTD-1k, the L1-GP-Rotated configuration achieves 69.19\% mIoU and $F_a=7.67\times10^{-6}$, outperforming SCTransNet (68.15\%, $F_a=10.74$) and all other baselines. On NUDT-SIRST, the same configuration attains 82.86\% mIoU, surpassing SCTransNet (80.69\%) by 2.17 points.
The diff-based (L1) loss consistently outperforms the Var-based (L2) counterpart: L1-GP-Rotated (69.19\%, $F_a=7.67$) vs.\ L2-GP-Rotated (68.56\%, $F_a=11.77$) on IRSTD-1k, confirming the benefit of strictly monotonic scale weighting. Mobius(L3) variants also show moderate and balanced performance across both datasets.
Gaussian-shaped spatial attention (GP) consistently improves mIoU and reduces $F_a$ across all scale loss families.

\begin{table}[t]
	\centering
	\caption{Comparison on IRSTD-1k, NUDT-SIRST, and SIRST-UAVB. IoU and $P_d$ are reported in percentage (\%), and $F_a$ is reported in $\times 10^{-6}$. Higher is better for IoU and $P_d$, while lower is better for $F_a$.}
	\label{tab:unified_compare_fused}
	\renewcommand{\arraystretch}{1.15}
	\small
	\resizebox{\textwidth}{!}{
		\begin{tabular}{l|ccc|ccc|ccc}
			\toprule
			\multirow{2}{*}{Methods}
			& \multicolumn{3}{c|}{IRSTD-1k}
			& \multicolumn{3}{c|}{NUDT-SIRST}
			& \multicolumn{3}{c}{SIRST-UAVB} \\
			\cmidrule(lr){2-4}\cmidrule(lr){5-7}\cmidrule(lr){8-10}
			& mIoU$\uparrow$ & $P_d\uparrow$ & $F_a\downarrow$
			& mIoU$\uparrow$ & $P_d\uparrow$ & $F_a\downarrow$
			& mIoU$\uparrow$ & $P_d\uparrow$ & $F_a\downarrow$ \\
			\midrule
			ACM          & 57.03 & 93.27 & 65.28  & 64.40 & 93.12 & 55.22 & 18.46 & 69.24 & 11.08\\
			DNANet       & 66.38 & 90.91 & 12.24  & 76.47 & 97.31 & 14.13 &23.96 &63.72  & 15.11  \\
			UIU-Net      & 66.66 & 93.98 & 22.07  & 78.09 & 96.77 & 12.54 & 26.84 & 66.57 & 7.38 \\
			MTU-Net      & 63.24 & 93.27 & 36.80  & 77.54 & 93.97 & 46.95 & 25.74 & 65.49 & 8.82 \\
			SCTransNet   & 68.15 & 93.27 & 10.74  & 80.69 & 95.71 & 16.52 & 27.33 & 74.34 & 8.87 \\
			\midrule
			\textbf{MSHNet (L1-GP-Rotated)} & \textbf{69.19} & 94.22 & \textbf{7.67}  & \textbf{82.86} & 97.40 & 20.89 & \textbf{27.91} & \textbf{73.90} & 8.85 \\
			MSHNet (L1-GP)                    & 68.98 & 93.20 & 9.95   & 82.55 & 97.42 & 16.16 & 23.58 & 64.50 & 8.87 \\
			MSHNet (L1)                       & 67.71 & 94.22 & 16.02  & 80.77 & 97.33 & 12.87 & 23.80 & 63.48 & 7.21 \\
			MSHNet (L1-ONLY-GP)               & 64.87 & 92.52 & 20.42  & 81.38 & 95.75 & 16.24 & 22.37 & 61.92 & 9.65 \\
			MSHNet (L2-GP-Rotated)          & 68.56 & 93.54 & 11.77  & 78.21 & \textbf{97.60} & 21.25 & 27.09 & 68.97 & \textbf{6.82} \\
			MSHNet (L2-GP)                    & 65.44 & 93.54 & 26.57  & 78.45 & 96.46 & \textbf{10.01} & 26.22 & 72.03 & 8.21 \\
			MSHNet (L2)                       & 67.34 & 93.85 & 15.03  & 79.23 & 98.84 & 11.64 & 24.00 & 66.42 & 8.98 \\
			MSHNet (L2-ONLY-GP)               & 65.30 & 93.20 & 23.76  & 77.39 & 97.20 & 22.27 & 20.70 & 60.72 & 10.76 \\
			MSHNet (L3-GP-Rotated)          & 67.45 & 92.86 & 11.24  & 81.44 & 96.30 & 10.68 & 26.89 & 69.84 & 8.48 \\
			MSHNet (L3-GP)                    & 67.18 & 92.19 & 11.46  & 82.74 & 97.53 & 11.74 & 26.65 & 70.56 & 9.09 \\
			MSHNet (L3)                       & 66.65 & 92.86 & 18.22  & 82.90 & 97.50 & 12.20 & 26.72 & 73.92 & 9.09 \\
			MSHNet (L3-ONLY-GP)               & 68.28 & \textbf{94.90} & 12.68  & 76.45 & 97.10 & 22.24 & 22.00 & 67.50 & 9.59 \\
			MSHNet (L4-GP-Rotated)          & 68.25 & 93.54 & 8.45   & 78.01 & 97.40 & 27.50 & 27.71 & 69.37 & 8.37 \\
			MSHNet (L4-GP)                    & 67.33 & 93.54 & 8.20   & 76.79 & 93.50 & 22.93 & 16.84 & 66.57 & 9.38 \\
			MSHNet (L4)                       & 65.78 & 91.50 & 17.69  & 79.95 & 96.28 & 12.55 & 25.82 & 72.00 & 10.54 \\
			MSHNet (L4-ONLY-GP)               & 64.79 & 90.82 & 24.82  & 73.09 & 97.00 & 38.68 & 15.95 & 52.65 & 9.21 \\
			\bottomrule
		\end{tabular}
	}
\end{table}
\subsection{Plug-and-play Comparisons}

\begin{figure}[t]
	\centering
	\includegraphics[width=1.0\textwidth]{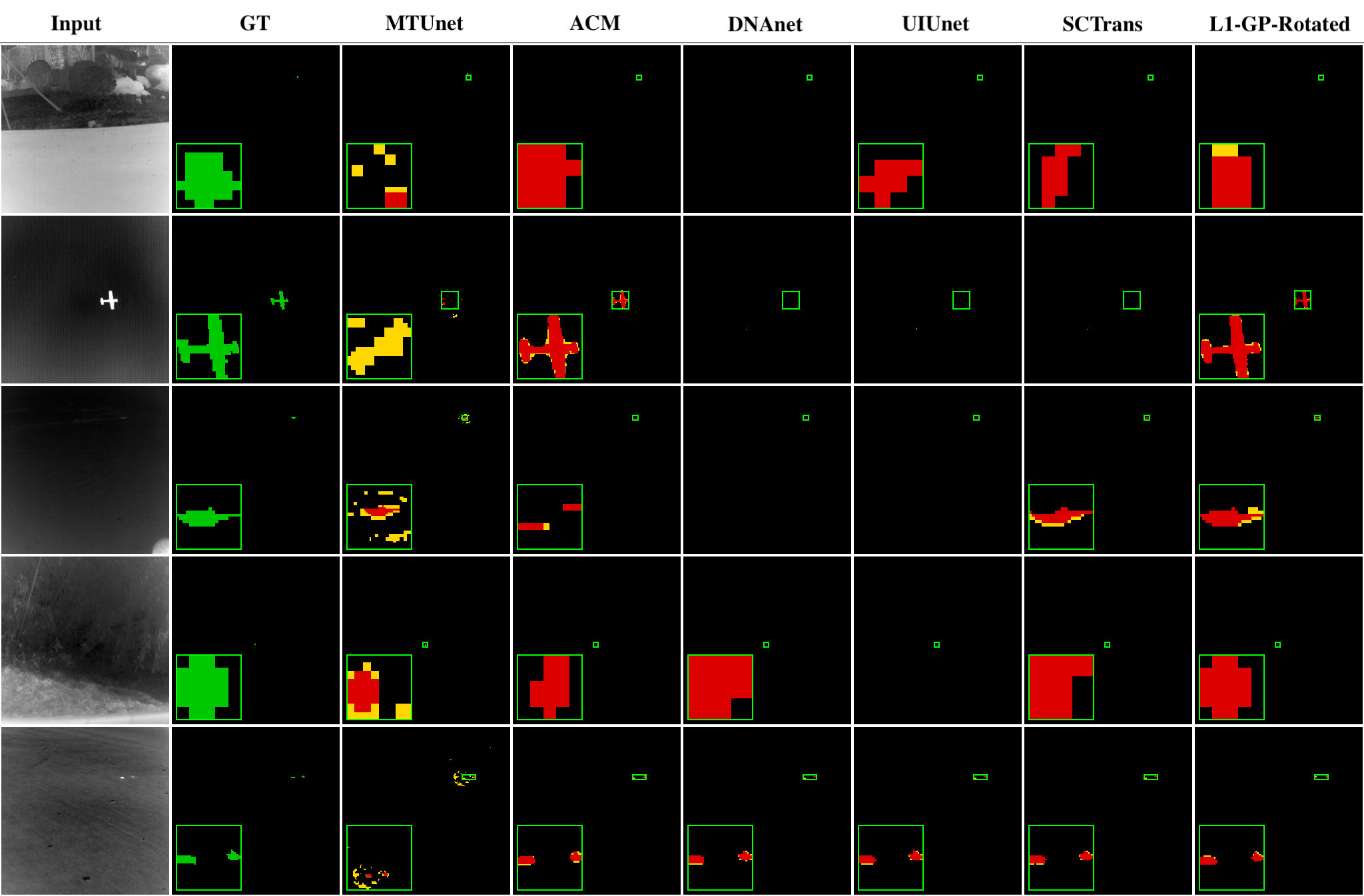}
	\caption{Qualitative segmentation comparison. Each row shows one test scene. Zoomed insets highlight the target region. Green pixels indicate true positives, red pixels indicate false positives, and yellow pixels indicate false negatives. L1-GP-Rotated consistently produces clean segmentation masks with fewer false alarms and missed detections compared to competing methods.}
	\label{fig:qualbox}
\end{figure}

\begin{figure}[t]
	\centering
	\includegraphics[width=1.0\textwidth]{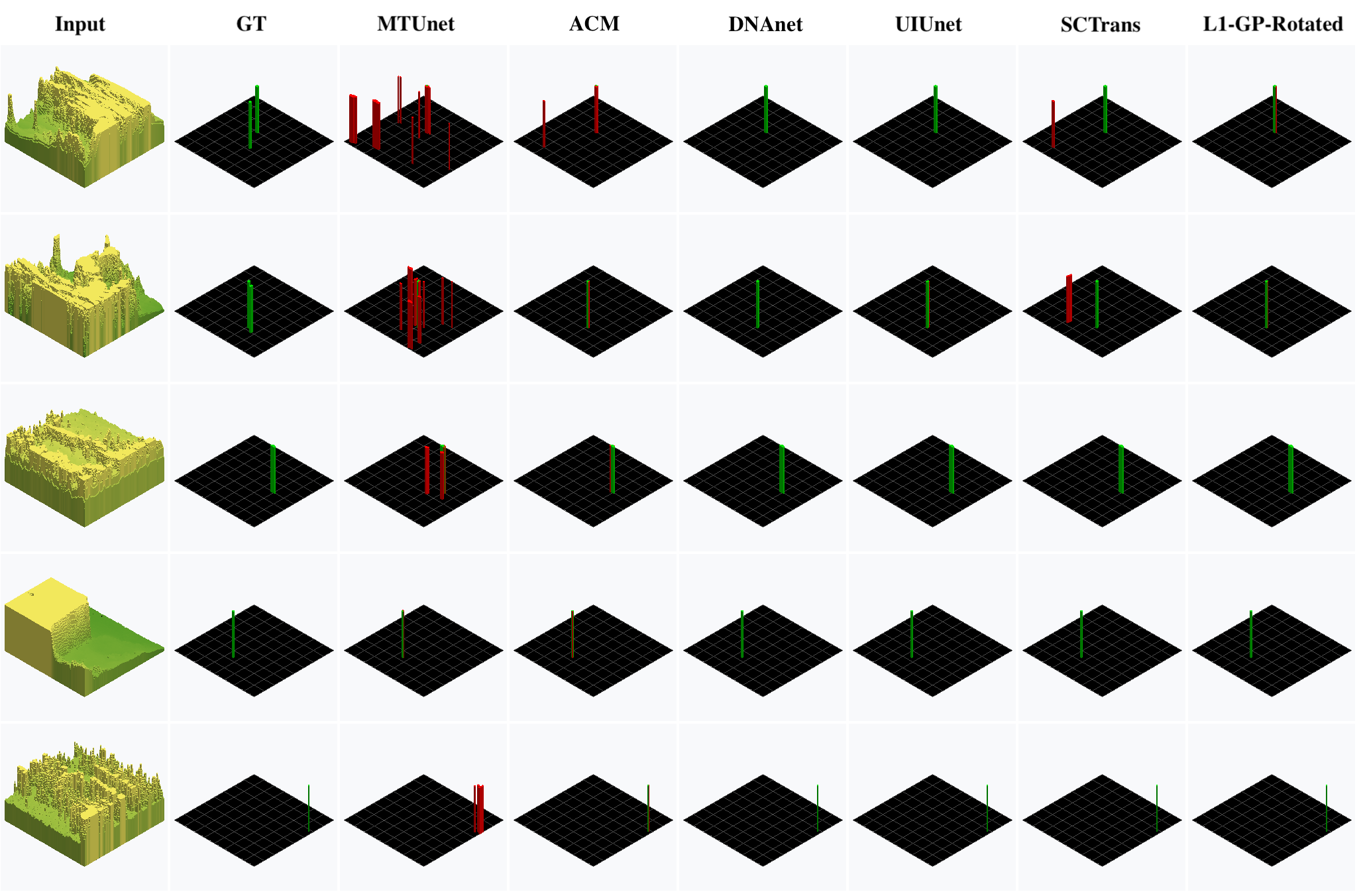}
	\caption{3D prediction heatmap comparison. Each bar represents a detected blob: green bars are true detections at the correct location, red bars are false alarms. L1-GP-Rotated yields a single clean green bar aligned with the ground-truth target, demonstrating significantly better false alarm suppression.}
	\label{fig:qualheat}
\end{figure}
 To assess whether the proposed Gaussian-shaped convolution with rotated mask also benefits \emph{detection}-oriented architectures, we conduct a plug-and-play study, using the standard convolution in the backbone of two generic detectors—YOLOv8n and RetinaNet—is replaced by GP-Rotated and seven alternative convolution designs, while all other components remain unchanged. We evaluate on IRSTD-1K and SIRST-UAVB using precision ($P$), recall ($R$), and mAP50. Table~\ref{tab:plugplay} reports the results.
GP-Rotated achieves the highest mAP50 on both datasets under YOLOv8n (88.5\% on IRSTD-1K, 95.1\% on SIRST-UAVB) and attains the best mAP50 under RetinaNet as well (71.1\% on IRSTD-1K, 89.5\% on SIRST-UAVB), while maintaining the same parameter count as standard convolution (3.048M). These results demonstrate that the Gaussian-shaped spatial prior is not only effective in segmentation but transfers directly to detection tasks, consistently improving detection accuracy for infrared small targets across different detector architectures without any additional parameters.

\begin{table}[t]
    \centering
    \caption{Plug-and-play comparison of convolution modules on YOLOv8n and RetinaNet across IRSTD-1K and SIRST-UAVB. $P$: precision (\%), $R$: recall (\%), mAP50 (\%), Params: total parameters (M).}
    \label{tab:plugplay}
    \renewcommand{\arraystretch}{1.1}
    \resizebox{\textwidth}{!}{
    \begin{tabular}{l|ccc|ccc|c|ccc|ccc}
        \toprule
        \multirow{3}{*}{Conv Module}
        & \multicolumn{7}{c|}{YOLOv8n Detection}
        & \multicolumn{6}{c}{RetinaNet Detection} \\
        \cmidrule(lr){2-8}\cmidrule(lr){9-14}
        & \multicolumn{3}{c|}{IRSTD-1K} & \multicolumn{3}{c|}{SIRST-UAVB} & \multirow{2}{*}{Params}
        & \multicolumn{3}{c|}{IRSTD-1K} & \multicolumn{3}{c}{SIRST-UAVB} \\
        \cmidrule(lr){2-4}\cmidrule(lr){5-7}\cmidrule(lr){9-11}\cmidrule(lr){12-14}
        & $P$ & $R$ & mAP50 & $P$ & $R$ & mAP50 & & $P$ & $R$ & mAP50 & $P$ & $R$ & mAP50 \\
        \midrule
        Conv     & 88.0 & 80.6 & 85.9 & 83.9 & 79.9 & 83.6 & 3.048 & 8.2  & 21.8 & 31.3 & 12.6 & 23.9 & 46.7 \\
        DySConv  & 87.9 & 79.4 & 85.8 & 87.7 & 83.7 & 88.1 & 3.117 & 23.5 & 35.3 & 66.7 & 36.2 & 48.2 & 84.4 \\
        DWConv   & 81.2 & 74.4 & 77.6 & 78.5 & 51.1 & 59.6 & 2.660 & 23.4 & 34.5 & 69.2 & 36.8 & 48.0 & 86.4 \\
        DSConv   & 79.8 & 75.7 & 80.6 & 90.6 & 92.3 & 94.3 & 2.796 & 23.4 & 35.0 & 70.1 & 37.1 & 48.1 & 86.5 \\
        WSConv   & 86.6 & \textbf{83.7} & 88.3 & 88.9 & 89.5 & 92.9 & 3.011 & 24.2 & 35.0 & 69.1 & 18.4 & 30.3 & 50.6 \\
        DConv    & 90.4 & 80.1 & 79.1 & 88.0 & 84.9 & 89.2 & 2.786 & \textbf{27.9} & 35.7 & 69.9 & 28.3 & 38.6 & 76.2 \\
        PConv    & 87.6 & 82.4 & 86.2 & 91.3 & 89.0 & 91.9 & 2.802 & 21.5 & 35.1 & 64.9 & 40.3 & 48.7 & 87.9 \\
        LDConv   & 89.5 & 81.2 & 86.1 & 89.6 & 89.2 & 92.7 & 2.791 & 24.1 & 35.2 & 67.9 & 40.4 & 49.2 & 87.9 \\
        \midrule
        GP-Rotated & \textbf{92.6} & 82.1 & \textbf{88.5} & \textbf{93.5} & \textbf{92.9} & \textbf{95.1} & 3.121 & 26.9 & \textbf{35.8} & \textbf{71.1} & \textbf{49.9} & \textbf{49.3} & \textbf{89.5} \\
        \bottomrule
    \end{tabular}
    }
\end{table}

\subsection{High False Alarm Source}

We examine whether the location regularizer $\mathcal{L}_{\text{loc}}$ is the structural cause of elevated $F_a$.
Table~\ref{tab:other_exp} compares four settings on IRSTD-1k and NUDT-SIRST.
The results provide two complementary pieces of evidence against the hypothesis that $\mathcal{L}_{\text{loc}}$ causes $F_a$ explosion.
First, \texttt{L1-ONLY-GP} removes $\mathcal{L}_{\text{loc}}$ entirely and retains only the scale loss; its $F_a$ remains at 20.42 on IRSTD-1k and 16.24 on NUDT-SIRST—neither lower nor dramatically higher than the full \texttt{L1-GP} configuration (9.95 / 16.16).
Taken together, the variation in $F_a$ across configurations is better explained by stochastic training dynamics and dataset-specific difficulty than by any structural side effect of the location regularizer.
\begin{table}[t]
    \centering
    \caption{Effect of location loss on $F_a$. mIoU and $P_d$ in \%; $F_a$ in $\times10^{-6}$.}
    \label{tab:other_exp}
    \renewcommand{\arraystretch}{1.1}
    \resizebox{\textwidth}{!}{
    \begin{tabular}{l|ccc|ccc}
        \toprule
        \multirow{2}{*}{Methods}
        & \multicolumn{3}{c|}{IRSTD-1k}
        & \multicolumn{3}{c}{NUDT-SIRST} \\
        \cmidrule(lr){2-4}\cmidrule(lr){5-7}
        & mIoU$\uparrow$ & $P_d\uparrow$ & $F_a\downarrow$
        & mIoU$\uparrow$ & $P_d\uparrow$ & $F_a\downarrow$ \\
        \midrule
        MSHNet (LLOSS-ONLY-GP) & 60.94 & 90.48 & 12.22 & 77.13 & 92.16 & 23.59 \\
        MSHNet (L1-ONLY-GP)    & 64.87 & 92.52 & 20.42 & 81.38 & 95.75 & 16.24 \\
        MSHNet (L1-GP)         & 68.98 & 93.20 & 9.95  & 82.55 & \textbf{97.42} & \textbf{16.16} \\
        MSHNet (L1-GP-Rotated) & \textbf{69.19} & \textbf{94.22} & \textbf{7.67} & \textbf{82.86} & 97.40 & 20.89 \\
        \bottomrule
    \end{tabular}
    }
\end{table}

\subsection{Qualitative Results}

Fig.~\ref{fig:qualbox} and Fig.~\ref{fig:qualheat} present qualitative comparisons of segmentation masks and 3D prediction heatmaps across five representative scenes on IRSTD-1k, against MTUnet, ACM, DNAnet, UIUnet, and SCTrans. The proposed L1-GP-Rotated configuration produces consistently cleaner segmentation outputs, target regions are accurately delineated with fewer false alarms and missed detections, even in challenging scenarios with cluttered backgrounds, low signal-to-noise ratio, or elongated target morphologies. In contrast, competing methods generate fragmented masks, over-segmented blobs, or entirely miss the target. The 3D heatmaps further confirm  the proposed method concentrates its response into a single bar precisely aligned with the ground-truth target location, demonstrating the effectiveness of the diff-based scale loss in suppressing false alarms and the Gaussian-shaped spatial attention in focusing detection on true target regions.

\section{Conclusion}

We propose a diff-based scale loss with strictly monotonic gradients eliminating training instability from the non-monotonic variance-based alternative and a Gaussian-shaped convolution with learnable $\sigma$ modulated by a learnable rotated pinwheel mask aligning the spatial receptive field with the physical imaging characteristics of small targets.
Experiments on IRSTD-1k, NUDT-SIRST, and SIRST-UAVB demonstrate consistent improvements in mIoU, $P_d$, and $F_a$, and plug-and-play evaluations on YOLOv8n and RetinaNet confirm the module's generality to detection architectures. 
\bibliographystyle{splncs04}
\bibliography{references}

\end{document}